\documentclass[11pt,a4paper]{article}
\usepackage[hyperref]{acl2020}
\usepackage{times}
\usepackage{latexsym}

\usepackage{microtype}

\aclfinalcopy 

\usepackage{booktabs}
\usepackage{graphicx}  
\usepackage{etex}
\usepackage{enumitem}
\usepackage{tikz-qtree-compat}
\usepackage{amsmath, amsfonts, amsthm, amssymb}
\usepackage{forest} 
\usepackage{latexsym}
\usepackage{relsize}
\usepackage{caption} 
\usepackage{colortbl} 
\usepackage{pifont}
\usepackage{subcaption} 
\usepackage{soul} 
\usepackage{tcolorbox} 
\usepackage{tikz}  
\usetikzlibrary{arrows} 
\usepackage{todonotes} 
\usepackage{mathalfa} 
\usepackage{mathtools}
\usepackage{mathrsfs}
\usepackage{multirow}
\usepackage{marvosym}
\usepackage[frozencache,cachedir=.]{minted}
\usepackage[e]{esvect}

\usepackage{bm}

\tcbuselibrary{skins} 
\usepackage{adjustbox}

\usepackage{naturaltableau}
\forestset{
  labA/.style n args=3{edge label/.expanded={node [midway, auto=#1, font=\unexpanded{#2}]{\unexpanded{#3}}},
  },
  labB/.style n args=3{label={[label distance=#1, font=\unexpanded{#2}]-90:{\unexpanded{#3}}},
  },
  labelA/.style ={edge label/.expanded={node [midway, auto=right, font=\unexpanded{\footnotesize}]{\unexpanded{#1}}},
  },
  labelB/.style ={label={[label distance=3pt, font=\unexpanded{\footnotesize}]-90:{\unexpanded{#1}}},
  }
}

\newcommand{\keycon}[1]{\emph{\textbf{#1}}} 

\newcommand{\sickd}[0]{\textsc{sick}}
\newcommand{\sickid}[1]{\textsc{sk}-#1}

\newcommand{\smtt}[1]{\texttt{\small #1}}
\newcommand{\cntxt}[1]{\textcolor{black!70}{\small\it#1}}

\newsavebox{\closureRule}

\title{Learning as Abduction:\\
\medskip
Trainable Natural Logic Theorem Prover for Natural Language Inference}

\author{Lasha Abzianidze
\\
  UiL OTS, Utrecht University\\
  \texttt{l.abzianidze@uu.nl}}

\date{}

\begin{document}
\maketitle
\begin{abstract}
  Tackling Natural Language Inference with a logic-based method is becoming less and less common.
  While this might have been counterintuitive several decades ago, nowadays it seems pretty obvious. 
  The main reasons for such a conception are that (a) logic-based methods are usually brittle when it comes to processing wide-coverage texts, and (b) instead of automatically learning from data, they require much of manual effort for development.
  We make a step towards to overcome such shortcomings by modeling learning from data as abduction: reversing a theorem-proving procedure to abduce semantic relations that serve as the \emph{best} explanation for the gold label of an inference problem.
  In other words, instead of proving sentence-level inference relations with the help of lexical relations, the lexical relations are \emph{proved} taking into account the sentence-level inference relations.
  We implement the learning method in a tableau theorem prover for natural language and show that it improves the performance of the theorem prover on the \sickd{} dataset by $1.4\%$ while still maintaining high precision ($>94\%$).
  The obtained results are competitive with the state of the art among logic-based systems.
\end{abstract}

\blfootnote{
\hspace{-0.65cm}
This work is licensed under a Creative Commons Attribution 4.0 International License. License details: \url{http://creativecommons.org/licenses/by/4.0/}.}

\section{Introduction}
\label{sect:introduction}

Natural language inference (NLI) is a well-established task for measuring intelligent systems' capacity of natural language understanding \cite{fracas,Dagan:2005}.
To improve and better evaluate the systems on the NLI task, many annotated NLI datasets are prepared and used for training and evaluating NLP models.
Generally speaking, an NLI dataset is a set of natural language sentence pairs, called premise-hypothesis pairs, that are annotated by crowd workers with one of three inference labels (\textit{entailment}, \textit{contradiction}, and \textit{neutral}), representing a semantic relation from a premise to a hypothesis.

Currently, the state-of-the-art systems in NLI are exclusively based on Deep Learning (DL).
One of the reasons for this is that DL-based systems can eagerly learn from task-related data and also take an advantage form high-quality pre-trained word embeddings.
The training phase helps them to obtain competitive scores on the in-domain test part.
On the other hand, logic-based systems are becoming less favored for NLI since it is hard to scale them up for reasoning with wide-coverage sentences.
Despite some rare exceptions \cite{martinez-gomez-etal-2017-demand,yanaka-etal-2018-acquisition}, it is notoriously hard to effectively and efficiently train logic-based systems on NLI datasets.

\bgroup
\setlength{\textfloatsep}{7pt plus 5.0pt minus 5.0pt}
\begin{figure}[t]
\hspace*{0mm}\includegraphics[clip, trim=0cm 4.7cm 9cm 0cm, 
    width=.49\textwidth]{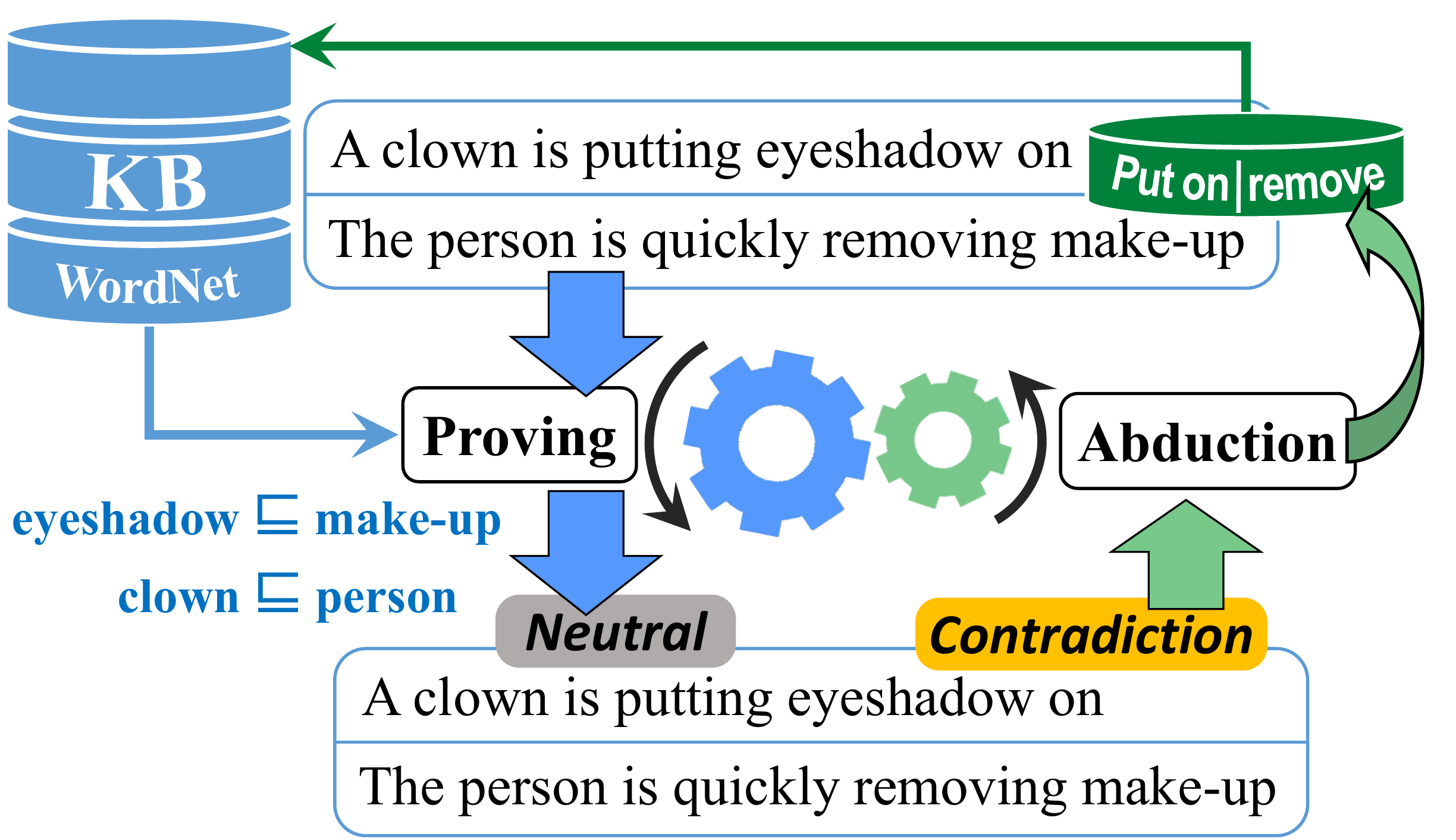}
\vspace{-6mm}
\caption{Theorem proving retrieves necessary semantic relations from KB, e.g., \emph{eyeshadow} is {make-up}.
To learn domain-specific relations, e.g., \emph{put on} is incompatible with {remove} (not found in KB), abduction reverse proves semantic relations from training samples.}
\label{fig:birdview}
\end{figure}
\egroup

Although DL-based systems are more robust than logic-based ones, the latter systems offer unique virtues such as a transparent reasoning procedure and reasoning with multiple premises.
An opaque decision procedure of DL-based systems makes it difficult to estimate a share of knowledge from what was learned by the systems, because not all what is learned is knowledge. 
Behind high performance of DL-based systems on particular NLI datasets, one might miss the systems' inability of generalization \cite{glockner-etal-2018-breaking} or the exploitation of annotation artefacts \cite{poliak-etal-2018-hypothesis,gururangan-etal-2018-annotation}.%

In this paper, we are not comparing logic- and DL-based approaches with respect to the NLI task.
Rather, we are proposing a learning method which demonstrates how a logic-based NLI system can be trained on NLI dataset, the aspect in which DL approaches to NLI significantly outperform symbolic approaches.
The proposed learning algorithm is inspired by abductive reasoning, which is often referred to as \emph{inference to the best explanation}.
Following the abduction, the algorithm allows learning those semantic relations over words and short phrases that \emph{best} explain gold inference labels of NLI problems (see \autoref{fig:birdview}).
In this way, the current work contributes to automated knowledge acquisition from data, which is considered as a major issue in NLI \cite[p.\,7]{rteBook:2013}.

The paper makes contributions along two lines: 
(a) describing how learning as abduction enables a trainable theorem prover for NLI, and
(b) implementing the algorithm and evaluating its effectiveness.
The original aspect of the research is the conceptual simplicity of the learning algorithm.
In particular, the standard workflow of the logic-based theorem prover is \emph{reversed}: instead of proving sentence-level inference relations with the help of lexical relations, the lexical relations are \emph{proved} taking into account the sentence-level inference relations.   
Throughout the paper we answer the following research questions:
\begin{enumerate}[label={\textsc{q}\arabic*}, topsep=0pt, itemsep=2pt, partopsep=2mm, parsep=0pt]
    \item What is a computationally feasible learning method that allows training the natural language theorem prover on NLI problems?
    \label{q1}
    \item How can learning pseudo-knowledge be avoided?
    \label{q2}
    \item Can the learned knowledge replace the lexical knowledge database like WordNet?
    \label{q3}
    \item To what extent the learned knowledge boosts the performance of the prover?
    \label{q4}
\end{enumerate}

The rest of the paper briefly introduces the natural language theorem prover (\autoref{sect:nltp}), 
describes the new learning algorithm motivated by abduction (\autoref{sect:alg}), 
outlines settings of experiments (\autoref{sect:exp}), 
reports and analyzes results of the experiments (\autoref{sect:res} and \autoref{sect:analysis}),
overviews related work and compares it with the current one (\autoref{sect:rel}), 
and finally, concludes the paper by answering the research questions (\autoref{sect:con}).

\section{Natural Language Theorem Prover}
\label{sect:nltp}

For our experiments, we employ a natural language theorem prover, called LangPro \cite{abzianidze-2017-langpro},
which is an implementation of Natural Tableau---an analytic tableau system for natural logic \cite{muskens:10,abzianidzethesis}.
An inference procedure is more central to Natural Tableau and its prover than it is usually for other logic-based NLI systems \cite{BosMarkert2005EMNLP,mineshima-etal-2015-higher}, which first derives meaning representations and then uses a proof engine for inference.
The inference in Natural Tableau not only helps to prove semantic relations but also further expands semantics of logical forms (e.g., shifting from higher to lower-order terms).
This makes it difficult to separate inference and semantic representations in Natural Tableau.
Its central role of inference makes LangPro a suitable candidate for the data-driven learning experiment based on automated theorem proving.

\bgroup
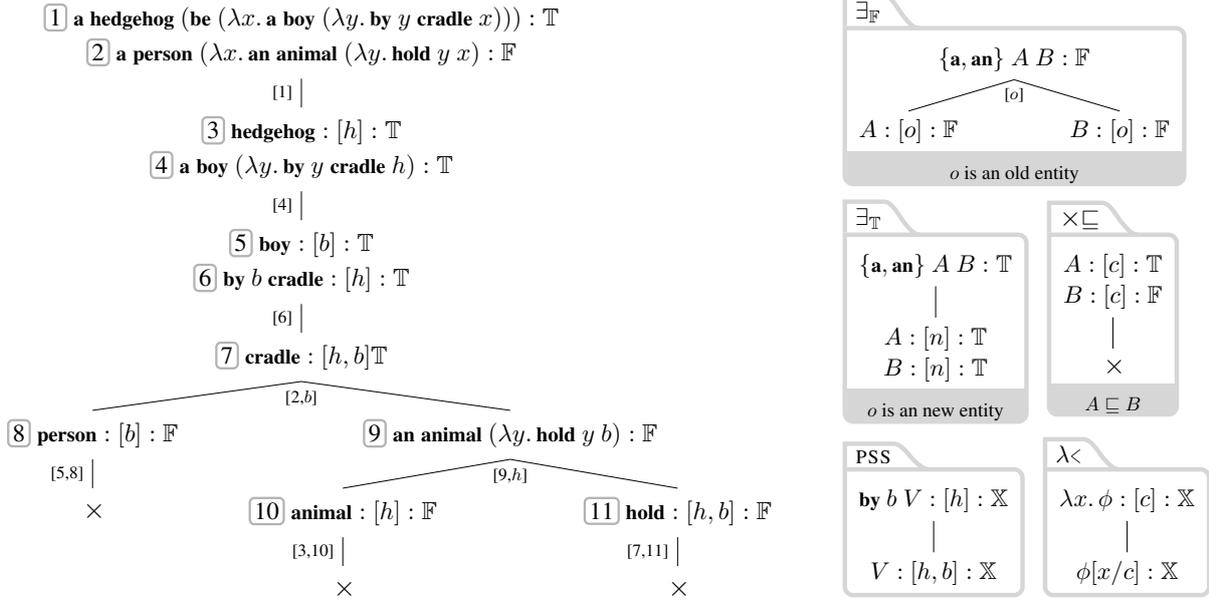
\begin{figure*}[t]
\noindent\begin{minipage}{\textwidth}
\begin{minipage}[t]{100mm}
\hspace*{-3mm}
\scalebox{.85}{\begin{forest}
baseline,for tree={align=center, parent anchor=south, child anchor=north, l sep=4.5mm, s sep=20mm}
[{\lab{1}~$\synt{a}~\synt{hedgehog}~(\synt{be}~(\lambda x.\,\synt{a}~\synt{boy}~(\lambda y.\,\synt{by}~y~\synt{cradle}~x))) : \T$}\\
 {\lab{2}~$\synt{a}~\synt{person}~(\lambda x.\,\synt{an}~\synt{animal}~(\lambda y.\,\synt{hold}~y~x) : \F$}\toppad{4mm}
 [{\lab{3}~$\synt{hedgehog} : [h] : \T$}\\
  {\lab{4}~$\synt{a}~\synt{boy}~(\lambda y.\,\synt{by}~y~\synt{cradle}~h) : \T$}\toppad{4mm},
  	labelA={\ndList{1}},
  [{\lab{5}~$\synt{boy} : [b] : \T$}\\
   {\lab{6}~$\synt{by}~b~\synt{cradle} : [h] : \T$}\toppad{4mm},
   labelA={\ndList{4}},
   [{\lab{7}~$\synt{cradle} : [h,b] \T$},
    labelA={\ndList{6}},
    labB={0mm}{\footnotesize}{\ndList{2,$b$}}
      [{\lab{8}~$\synt{person} : [b] : \F$}
       [{$\btimes$},
      	  labelA={\ndList{5,8}}            
       ]
      ]
      [{\lab{9}~$\synt{an}~\synt{animal}~(\lambda y.\,\synt{hold}~y~b) : \F$},
         labB={0mm}{\footnotesize}{\ndList{9,$h$}}
       [{\lab{10}~$\synt{animal} : [h] : \F$}
         [{$\btimes$},
           labelA={\ndList{3,10}} 
         ]
       ]  
       [{\lab{11}~$\synt{hold} : [h,b] : \F$}
        [{$\btimes$},
           labelA={\ndList{7,11}}            
        ]
       ] 
      ]
   ]      
  ]      
 ]       
]
\end{forest}}
\end{minipage}%
\hspace{7mm}
\scalebox{.83}{
\begin{minipage}[t]{60mm}
\renewcommand{\arraystretch}{0}
\tabRuleFrame{$\exists_\F$}{\begin{forest}
for tree={align=center, parent anchor=south, child anchor=north, l sep=5mm, s sep=15mm}
[{$\{\synt{a}, \synt{an}\}~A~B : \F$},labB={0mm}{\footnotesize}{\ndList{$o$}}
    [{$A : [o] : \F$}]
    [{$B : [o] : \F$}]]
\end{forest}}{$o$ is an old entity}
\medskip

\tabRuleFrame{$\exists_\T$}{\begin{forest}
for tree={align=center, parent anchor=south, child anchor=north, l sep=5mm, s sep=15mm}
[{$\{\synt{a}, \synt{an}\}~A~B : \T$}
    [{$A : [n] : \T$}\\{$B : [n] : \T$}\toppad{4mm}]]
\end{forest}}{$o$ is an new entity}
~
\tabRuleFrame{\clSubs}{\begin{forest}
for tree={align=center, parent anchor=south, child anchor=north, l sep=5mm, s sep=15mm}
[{$A : [c] : \T$}\\{$B : [c] : \F$}\toppad{4mm}
    [{$\btimes$}]]
\end{forest}}{$A \sqsubseteq B$}
\medskip

\tabRuleFrame{\rulen{pss}}{\begin{forest}
for tree={align=center, parent anchor=south, child anchor=north, l sep=5mm, s sep=15mm}
[{$\synt{by}~b~V : [h] : \X$}
    [{$V : [h,b] : \X$}]]
\end{forest}}{}
~
\tabRuleFrame{\rulen{$\abstPull$}}{\begin{forest}
for tree={align=center, parent anchor=south, child anchor=north, l sep=5mm, s sep=15mm}
[{$\lambda x.\, \phi : [c] : \X$}
    [{$\phi[x/c] : \X$}]]
\end{forest}}{}
\end{minipage}
}

\end{minipage}
\vspace{-1mm}
\caption{On the left, a closed tableau which proves the entailment relation by failing to refute it. 
On the right, a set of inference rules that help to unfold semantics of larger terms.
}
\label{proof:demo_tree}
\end{figure*}
\egroup

The logic behind Natural Tableau and the prover is a \emph{higher-order logic} (aka \emph{simple type theory}) which also acts as a version of \emph{natural logic} \cite{Benthem:NatLog:2008,Moss2010}.
The $\lambda$-terms, given below with their corresponding sentences, represent logical forms of the natural logic.
\bgroup
\setlength{\abovedisplayskip}{4pt}
\setlength{\belowdisplayskip}{4pt}
\setlength{\abovedisplayshortskip}{0pt}
\setlength{\belowdisplayshortskip}{0pt}
\begin{align}
& \text{A hedgehog is cradled by a boy}
\label{snt:cradled}
\\[-2pt]
& \synt{a}~\synt{hedgehog}\,\big(\synt{be}\,(\lambda x.\,\synt{a}~\synt{boy}\,(\lambda y.\,\synt{by}\,y\,\synt{cradle}\,x))\big)\kern-3mm
\tag{\arabic{equation}a} \label{llf:cradled}
\\
& \text{A person holds an animal}
\label{snt:hold}
\\[-2pt]
& \synt{a}~\synt{person}~(\lambda x.\,\synt{an}~\synt{animal}~(\lambda y.\,\synt{hold}~y~x))
\tag{\arabic{equation}a} \label{llf:hold}
\\
& \text{Most dogs which jumped also barked loud}
\label{snt:bark_often}
\\[-2pt]
& \synt{most}~(\synt{which}~\synt{jump}~\synt{dog})~\big(\synt{also}~(\synt{loud}~\synt{bark})\big)
\tag{\arabic{equation}a} \label{llf:bark_often}
\end{align}
\egroup
In addition to the lexical terms, the terms employ only variables and constants.
Therefore, common logical connectives (e.g., $\wedge$, $\neg$) and quantifiers (e.g., $\exists$, $\forall$) are not part of the formal language. 
The terms are built using the $\lambda$ abstraction and function application.%
\footnote{In formal semantics literature, the function application is often denoted as $@$, but for better readability, we omit it.
The function application is left-associative, e.g.,  $ABC = (AB)C$.
To keep the terms leaner, we hide typing information of lexical terms, like \synt{which} being of type $\big((\np\!\to\!\sen)\!\to\!\nou\big)\!\to\!\nou$.
}
The role of variables and $\lambda$ is to access and fill certain argument positions and control scope. 
Variables and $\lambda$ are mainly used for terms with arity two or more, like \synt{cradle} and \synt{hold}.
\newcite{abzianidze-2015-tableau} showed that the terms can be automatically obtained from the derivations of Combinatory Categorial Grammar (CCG, \citealp{Steedman:2000}).
%
%
Since the workflow for the theorem proving is important to understand the proposed learning algorithm, 
we demonstrate on the example how inference problems are solved by the prover. 

After the sentences of an NLI problem are parsed with a parser and converted into $\lambda$-terms, the \emph{natural tableau prover} verifies the problem on entailment and contradiction relations.
For example, to prove that \eqref{snt:cradled} entails \eqref{snt:hold}, the tableau prover searches for a situation that makes \eqref{llf:cradled} true and \eqref{llf:hold} false.
In other words, it attempts to build a counterexample model that refutes the entailment relation.
In \autoref{proof:demo_tree}, the proof tree, so-called tableau, depicts the search for the counterexample.

The tableau starts with \eqref{llf:cradled} being true and \eqref{llf:hold} false, expressed by the entries \lab{1} and \lab{2}.
But what are the meanings of \lab{1} and \lab{2}?
To flesh out their meanings, inference rules are applied to the entries.
In particular, \lab{1} produces \lab{3} and \lab{4} with the help of the rule ($\exists_\T$), which says: if \emph{an A does B} is true, then \emph{there is some entity, let's name it $n$, which is A and does B}.%
\footnote{Obtaining \lab{4} from \lab{1} additionally requires application of ($\abstPull$) and the rule for auxiliaries that will treat \synt{be} as an identity function and discard it.}
Further applying ($\exists_\T$) to \lab{4} introduces \lab{5} and \lab{6}.
\lab{7} is obtained from \lab{6} with the help of (\rulen{pss}), which paraphrases passive constructions with any truth sign ($\X$) as active constructions.
So far, the prover managed to fold out semantics of \lab{1}: there are $b$ and $h$ who are a boy \lab{5} and a hedgehog \lab{3}, respectively, and $b$ cradles $h$ \lab{7}.

Now it is a proper time to note that a branch in a proof tree represents a set of situations/models, and entries sitting on the branch describe corresponding situations.  
Therefore, for now, a single set of situations is built such that in all the situations a boy cradles a hedgehog, and \lab{2} is false.
To decompose the meaning of \lab{2}, ($\exists_\F$) is applied to it.
This splits the set of situations into two parts, the situations where $b$ is not a person and the situations where \lab{9} holds.%
\footnote{In the latter set of situations $b$ is also a person. 
This is not explicitly asserted in the tableau because it is redundant due to the completeness of the set of first-order logic tableau rules.} 
The situations of the left branch don't make sense as (\clSubs) detects that in those situations $b$ is a boy \lab{5} but not a person \lab{8}.
The situations of the right branch are further categorized when applying ($\exists_\F$) to \lab{9}.
As a result, both groups of situations are inconsistent as in one group $h$ is a hedgehog but not an animal, and in another group, \emph{cradle} relation is not \emph{hold} relation.
In the end, all the branches are closed, i.e., the tableau is closed, as they model inconsistent situations.
This means that the refutation attempt has failed, and there is no counterexample for the entailment relation.
Hence, it is proved that \eqref{llf:cradled} entails \eqref{llf:hold}, and accordingly \eqref{snt:cradled} entails \eqref{snt:hold}.

In principle, it is also necessary to verify the NLI problem for contradiction.
In that case, the tableau proof starts with \lab{1} and \lab{2} being true.
If neither entailment nor contradiction is proved, the problem is classified as neutral.

\section{Learning as Abduction}
\label{sect:alg}

\subsection{What to learn?}
\label{subsect:what_to_learn}

The tableau proof in \autoref{proof:demo_tree} illustrated how inference over sentences is reduced to the semantic relations over lexical items.
Namely, to prove that \eqref{snt:cradled} entails \eqref{snt:hold}, the prover needs to know: $\synt{boy}\subseq{}\synt{person}$, $\synt{hedgehog}\subseq{}\synt{animal}$, and $\synt{cradle}\subseq{}\synt{hold}$.
One could employ existing (lexical) knowledge resources as a reply to the need for lexical relations, but it is well known that such resources are never enough.
Compared to NLI systems with learning algorithms, logic-based NLI systems are much more vulnerable when it comes to the knowledge sparsity because a small, missing piece of knowledge can corrupt the entire reasoning process and the judgment.

While knowledge resources are still valuable for reasoning, learning from data is a crucial component of success when it comes to evaluation against large datasets.
In the tableau prover, two components directly contribute to the proof procedure: an inventory of rules (IR) and a set of relations (aka facts), called knowledge base (KB).
In principle, the distinction between inference rules and relations is not straightforward.
That's why we hereafter adopt the distinction between the inference rules and the facts of the KB as it is done in the tableau prover.
Some approaches might consider $\synt{boy}\subseq{}\synt{person}$ relation as an inference rule like $\forall x. \synt{boy}(x)\!\to\!\synt{person}(x)$, but following Natural Tableau, hereafter they will be considered as facts of KB.
In general, a rule is schematic and can rewrite entries that match its antecedent entries (and even allow branching that acts as disjunction, see \autoref{proof:demo_tree}).
On the other hand, relations in KB are fully lexicalized and have the form of $A\subseq{}B$ or $A\disj{}B$, where $A$ and $B$ are atomic or compound terms.
In this way, apart from $\synt{clean}\disj{}\synt{dirty}$ and $\synt{chop}\subseq{}\synt{cut}$, the relations like $\synt{webcam}\subseq{}\synt{digital~camera}$ and $\synt{lie down}\disj{}\synt{run away}$ are also considered as facts despite corresponding to relations over short phrases.

Both IR and KB would benefit from data-driven learning.
Learning new rules can be as important as learning new relations.
However, as an initial step, we find learning relations more feasible and effective than learning rules for two reasons:
(a) relations are fully-specified unlike the rules, and
(b) given that relations include phrases too, learning relations could compensate rules to a large extent.
Moreover, results from the relation learning can provide further insight into learning rules.
For instance, many relations with the terms of similar structure can provide evidence for creating a rule.

\begin{figure*}
\hspace*{-6mm}\scalebox{.8}{\begin{forest}
baseline,for tree={align=center, parent anchor=south, child anchor=north, l sep=5.5mm, s sep=1mm}
[{\lab{1}~$\synt{a}~\synt{hedgehog}~(\synt{be}~(\lambda x.\,\synt{a}~\synt{boy}~(\lambda y.\,\synt{by}~y~\synt{cradle}~x))) : \T$}\\
 {\lab{2}~$\synt{a}~(\bluetabhl{\synt{young}~\synt{person}})~(\lambda x.\,\synt{a}~(\bluetabhl{\synt{small}~\synt{animal}})~(\lambda y.\,\synt{hold}~y~x) : \F$}\toppad{4mm}
 [{\lab{3}~$\synt{hedgehog} : [h] : \T$}\\
  {\lab{4}~$\synt{a}~\synt{boy}~(\lambda y.\,\synt{by}~y~\synt{cradle}~h) : \T$}\toppad{4mm},
  	labelA={\ndList{1}},
  [{\lab{5}~$\synt{boy} : [b] : \T$}\\
   {\lab{6}~$\synt{by}~b~\synt{cradle} : [h] : \T$}\toppad{4mm},
   labelA={\ndList{4}},
   [{\lab{7}~$\synt{cradle} : [h,b] : \T$},
    labelA={\ndList{6}},
    labB={0mm}{\footnotesize}{\ndList{2,$b$}}
      [{\lab{8}~$\synt{young}~\synt{person} : [b] : \F$},
       labB={0mm}{\footnotesize}{\ndList{8}}
       [{\lab{12}~$\synt{young} : [b] : \F$}
        [\redtabhl{Open branch 1}]
       ]
       [{\lab{13}~$\synt{person} : [b] : \F$}
        [{$\btimes$},
      	  labelA={\ndList{5,13}}            
        ]
       ]
      ]
      [{\lab{9}~$\synt{a}~(\synt{small}~\synt{animal})~(\lambda y.\,\synt{hold}~y~b) : \F$},
         labB={0mm}{\footnotesize}{\ndList{9,$h$}}
       [{\lab{10}~$\synt{small}~\synt{animal} : [h] : \F$},
        labB={0mm}{\footnotesize}{\ndList{10}}
         [{\lab{14}~$\synt{small} : [h] : \F$}
          [\redtabhl{Open branch 2}]
         ]
         [{\lab{15}~$\synt{animal} : [h] : \F$}
          [{$\btimes$},
      	   labelA={\ndList{3,15}}            
          ]
         ]
       ]  
       [{\lab{11}~$\synt{hold} : [h,b] : \F$}
        [{$\btimes$},
           labelA={\ndList{7,11}}            
        ]
       ] 
      ]
   ]      
  ]      
 ]       
]
\end{forest}}
\kern-19mm
%
\put(-258,-95){\scalebox{.8}{
\begin{minipage}{45mm}
\tabRuleFrame{\clSubs\vphantom{$|$}}{\begin{forest}
for tree={align=center, parent anchor=south, child anchor=north, l sep=3mm, s sep=15mm}
[{$A : [c] : \T$}\\{$B : [c] : \F$}\toppad{4mm}
    [{$\btimes$\vspace{-8mm}}]]
\end{forest}}{$A \sqsubseteq B$}
~
\tabRuleFrame{\cldisj}{\begin{forest}
for tree={align=center, parent anchor=south, child anchor=north, l sep=2.7mm, s sep=15mm}
[{$A : [c] : \T$}\\{$B : [c] : \T$}\toppad{4mm}
    [{$\btimes$\vspace{-8mm}}]]
\end{forest}}{$A | B$}
\end{minipage}
}}
%
\put(-15,0){\scalebox{.77}{\begin{forest}
baseline,for tree={align=center, parent anchor=south, child anchor=north, l sep=1.5mm, s sep=6mm}
[\redtabhl{Open branch 1}\\
\lab{1}\\
\lab{2}\\
\lab{3}\\
\lab{4}\\
\lab{5}\\
\lab{6}\\
\lab{7}\\
\lab{8}\\
\lab{12}\\\toppad{4mm}
  $b_1^1: \{ \synt{boy} \subs \synt{young person} \}$\\\toppad{4mm}
  $b_2^1: \{ \synt{boy} \subs \synt{young} \}$\\\toppad{4mm}
  \st{$b_3^1: \{ \synt{hedgehog} \,|\, \synt{by}\,b~\synt{cradle} \}$}
]
\end{forest}}}
\put(85,0){\scalebox{.77}{\begin{forest}
baseline,for tree={align=center, parent anchor=south, child anchor=north, l sep=3.3mm, s sep=6mm}
[\redtabhl{Open branch 2}\\
\lab{1}\\
\lab{2}\\
\lab{3}\\
\lab{4}\\
\lab{5}\\
\lab{6}\\
\lab{7}\\
\lab{9}\\
\lab{10}\\
\lab{14}\\\toppad{4mm}
  $b_1^2: \{ \synt{hedgehog} \subs \synt{small animal} \}$\\\toppad{4mm}
  $b_2^2: \{ \synt{hedgehog} \subs \synt{small} \}$\\\toppad{4mm}
  \st{$b_3^2: \{ \synt{hedgehog} \,|\, \synt{by}\,b~\synt{cradle} \}$}\\\toppad{4mm}
  \st{$b_4^2: \{ \synt{by}\,b~\synt{cradle} \subs \synt{small} \}$}\\\toppad{4mm}
  \st{$b_5^2: \{ \synt{by}\,b~\synt{cradle} \subs \synt{small~animal} \}$}
]
\end{forest}}}
%
\put(62,-188){\scalebox{.75}{
\textbf{T-sets} $ = \Bigg\{ \bigcup\limits_i \cup B^i 
~\Bigg|~ \parbox{23mm}{$B^1 \subseteq \{b^1_1, b_2^1\}$\\\toppad{5mm}$B^2 \subseteq \{b_1^2,b^2_2\}$} \Bigg\} $}}
\begin{lrbox}{\closureRule}
\begin{minipage}{90mm}
\begin{minted}[bgcolor=black!5,framerule=.5pt,rulecolor=black!50,frame=single]{prolog}
cl_rule(sub, [Node1,Node2], Rels) :- 
    Node1 = (Term1, Args, true),
    Node2 = (Term2, Args, false),
    isa(Term1, Term2, Rels).
\end{minted}
\end{minipage}\kern-40mm
\end{lrbox}
\put(20,-226){\scalebox{0.7}{\usebox{\closureRule}}}
\vspace{-2mm}
\caption{An open tableau represents a failed attempt to prove the entailment.
Each open branch can be closed by three B-sets using (\clSubs) and (\cldisj) rules.
B-sets for $k$th open branch are generated by uniting basis sets $b_{i\ldots j}^k$ of the branch.
Basis sets are obtained by applying closure rules backwards.
A union of B-sets of all open branches forms a T-set which is sufficient knowledge to close the tableau and solve the NLI problem.
}
\label{proof:abd_tree}
\end{figure*}
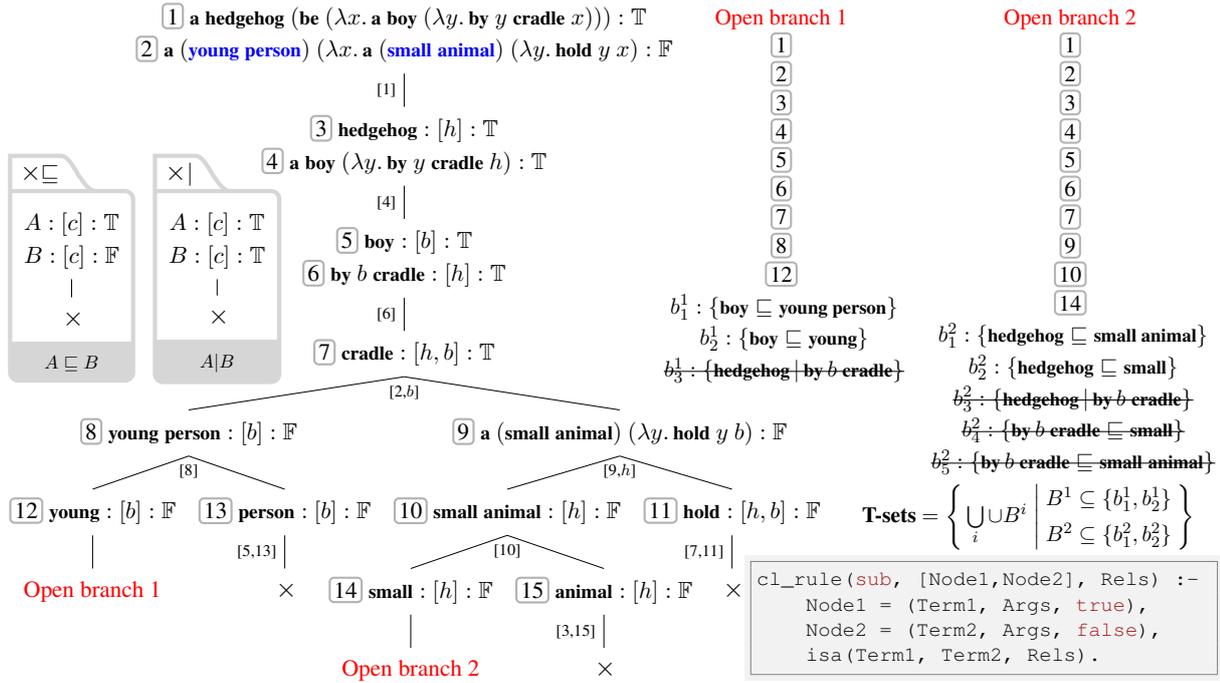

\subsection{Abductive Reasoning and Learning}
\label{subsect:abduction}

After deciding to learn relations for KB, the next step is to design a learning algorithm that takes a set of labeled NLI problems and produces a set of relations that boost the performance of the prover.
From a perspective of the tableau method, such relations should help to close corresponding tableau trees for the problems with entailment and contradiction labels.
Put differently, we search for relations explaining gold labels of NLI problems.

We formally define tableau-based abductive reasoning for NLI as a search problem:
given a labeled NLI problem $\mathcal{L}(P, H)$ and an optional KB denoted as $K$, find an explanation set of relations $E$ such that $\mathcal{L}(K \wedge E \wedge P, H)$ is provable.
Additionally, $E$ has to satisfy at least the following conditions:
\begin{enumerate}[label={A\arabic*}, leftmargin=10mm, listparindent=2mm, topsep=1mm, itemsep=0pt, partopsep=3mm, parsep=0pt, labelsep=2mm]
    \item $E$ is consistent with $K$;
    \label{A1}
    \item $E$ is minimal in the sense that $E \subseteq E'$ if $\mathcal{L}(K \wedge E' \wedge P, H)$ is provable;
    \label{A2}
    \item Relations in $E$ have some restricted form.
    \label{A3}
\end{enumerate} 
These conditions serve as minimal criteria for the concept of \emph{best} when assessing the explanations \cite{mayer_pirri:93}.

To illustrate abductive reasoning, let's consider an altered version of the example from \autoref{sect:nltp} where \emph{person} and \emph{animal} are modified. 
The alternation is such that it prevents the prover from finding a proof for the entailment relation.
The corresponding open tableau is given in \autoref{proof:abd_tree}.
The tableau is saturated, meaning that all possible rule applications were made while growing the tree.
Given that entailment is the gold label for the NLI problem, we need additional relations to close all the branches since $\synt{hedgehog}\subseq{}\synt{animal}$, $\synt{boy}\subseq{}\synt{person}$, and $\synt{cradle}\subseq{}\synt{hold}$ are not sufficient anymore.

The search for a set of relations that closes all open branches can be seen as searching for antecedent nodes of the closure rules, e.g., (\clSubs) and (\cldisj), and learning the constraint relations of the rules, e.g., $A\subs B$ and $A\disj{}B$, respectively.
Learning the relations enabling certain nodes to close a branch represents backwards application of closure rules.
This way of extracting relations suits well to LangPro as it is implemented in Prolog, which can be run forwards and backwards.
While for rule application, a set of relations \smtt{Rels} is specified in \smtt{cl\_rule/3} (see \autoref{proof:abd_tree}), during abductive learning \smtt{Rels} is initially unspecified and later specified by \smtt{isa/3}.
For example, when $\smtt{Node1}\!=\!\lab{3}$ and $\smtt{Node2}\!=\!\lab{10}$,  \smtt{cl\_rule/3} fails if \smtt{Rels} is specified and doesn't contain $\synt{hedgehog}\subseq{}\synt{small animal}$, but if \smtt{Rels} is unspecified, \smtt{cl\_rule/3} succeeds and \smtt{Rels} becomes a list containing $\synt{hedgehog}\subseq{}\synt{small animal}$.%

In the running example (\autoref{proof:abd_tree}), when considering only two closure rules (\clSubs) and (\cldisj), there are several sets of relations that close an open branch.
Let's call such a set of relations a \keycon{\mbox{B-set}}.
So, a B-set is specific to an open branch and closes it.
For example, a union of any non-empty subset of $\{b_1^1, b_2^1\}$ is a B-set for the first open branch.
The sets $b_1^1$ and $b_2^1$ are a \keycon{basis} of the B-sets of the first open branch, i.e., minimal sets that generate all B-sets of the branch.
The same applies to $b_1^2$ and $b_2^2$ for the second open branch.
Note that basis sets are automatically B-sets.
$b_3^1$ and $b^2_{3,4,5}$ sets are not B-sets as some of the terms in the relations are not fully lexicalized.
Let's call a set of relations a \keycon{T-set} if they help to close an entire tableau, i.e., close all open branches.
Therefore T-sets are potential explanations for the NLI problem. 
For instance, $\{ \synt{boy}\subseq{}\synt{young}, \synt{hedgehog}\subseq{}\synt{small}\}$ is one of the nine T-sets for the tableau.
The largest T-set is a union of all the basis B-sets.
To learn the \emph{best} T-set from the possible options, the next section presents criteria used to define the notion of \emph{the best}.

\subsection{Searching for \emph{the best}}
\label{subsect:best}

The tableau proof presented in \autoref{sect:nltp} is a toy example compared to the actual proof tree produced by the theorem prover which consists of 53 entries distributed over 8 branches.%
\footnote{The reason for the increase is that not all rule applications are relevant for the final proof, but this is impossible to anticipate beforehand.
}
This means that during the abduction, there will be more branches and more nodes per branch than in \autoref{proof:demo_tree}.
Also, taking into account all 16 closure rules of the prover, this amounts to a large number of B-sets and T-sets.
A set of all T-sets will serve as a search space for explanations in abductive reasoning.
To make the reasoning efficient, we filter out certain T-sets following (\ref{A1}-\ref{A3}) conditions.

First, according to \eqref{A3}, we decrease the number of possible T-sets by allowing only relations with \textit{\textbf{term shapes}} of $A, AB, (AB)C, A(BC)$, where each meta-variable is a lexical term.
In this way, T-sets with relations over terms of size four, like $(\synt{and}\,\synt{big}\,\synt{brown})\,\synt{dog}$, will be ignored.

To further narrow down types of learned relations along the lines of \eqref{A3}, we consider relations over \keycon{syntactically comparable terms}, where possible categories for open class words are noun, verb or adjective/adverb.%
\footnote{For each term, a head and a syntactic category can be detected using POS tags and CCG categories, which are assigned by a CCG parser and kept in the term representation.
}
This means that relations like $\synt{boy}\disj{}\synt{run}$ and $\synt{boy}\subseq{}\synt{young}$ will be ignored while keeping relations like $\synt{boy}\disj{}\synt{hedgehog}$ and $\synt{boy}\subseq{}\synt{young person}$.
We opt for this restriction because in-category semantic relations tend to be more  genuine than cross-category relations.
So, we expect learned in-category relations to generalize better in different contexts.

One of the criteria for the best explanation is minimality \eqref{A2}.
We interpret this as an \keycon{amount of information} and prefer minimal T-sets in terms of set inclusion to larger ones.
Therefore, the amount of information induces a partial ordering over T-sets.
Candidates for minimal T-sets can be formed by uniting minimal B-sets per open branch, where basis sets, e.g., $b_i^j$, are minimal B-sets.
For example, minimal T-sets for the tableau in \autoref{proof:abd_tree} are $b_1^1 \cup b_1^2$, $b_1^1 \cup b_2^2$, $b_2^1 \cup b_1^2$, and $b_2^1 \cup b_2^2$.
The intuition behind the information criterion is to learn as few relations as possible sufficient for proving an NLI problem and, hopefully, to prevent overfitting during the training.

\begin{figure*}[t]
\centering
\mbox{\includegraphics[clip, trim=0mm 92mm 3mm .9mm, 
    width=.85\textwidth]{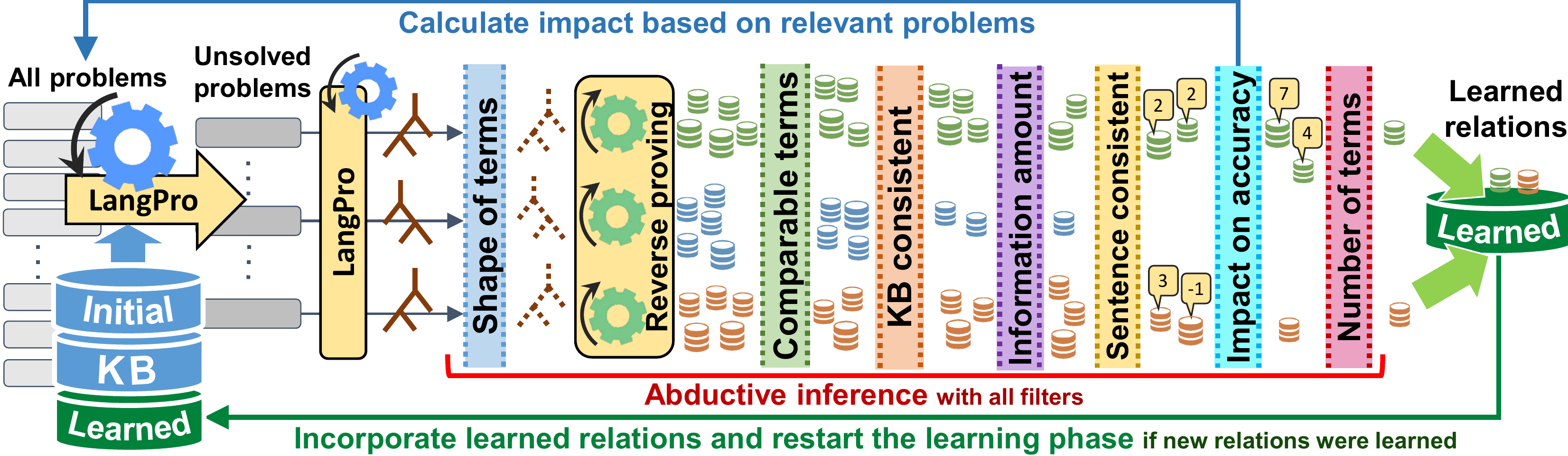}}
\vspace{-2mm}
\caption{Learning starts with an initial KB.
Abduction is carried out on unsolved entailment and contradiction problems.
Inferred knowledge, i.e., T-sets, pass several filters to select the \emph{best} knowledge.
The learned knowledge is added to the initial KB, and the learning phase repeats until no new knowledge is learned.} 
\label{fig:learning_pipe}
\end{figure*}

Following \eqref{A1}, a relation has to be \keycon{semantically consistent with existing KB}. In this way, relations like $\synt{hedgehog}\disj{}\synt{animal}$ or $\synt{big}\subseq{}\synt{small}$ will be dubbed inconsistent with KB which includes $\synt{hedgehog}\subseq{}\synt{animal}$ and $\synt{big}\disj{}\synt{small}$.
In experiments, instead of doing a complete consistency checking of the entire KB every time a new relation is considered, we perform a lazy check by verifying whether $A\disj{}B$ and one of $A\subseq{}B$ and $B\subseq{}A$ together occur in KB.
We go further and ignore relations of form $B\disj{}AB$ taking into account that subsective lexical modifiers are prevalent.
Hence, relations like $\synt{small animal}\disj{}\synt{animal}$ will be dropped.

Additionally, we consider only such T-sets that are \keycon{semantically consistent with sentences} of the corresponding NLI problem.  
This can be seen as further elaboration on \eqref{A2} since the sentences are usually consistent with background knowledge.
The example of a bad T-set, inconsistent with the sentence, is the one containing $\synt{baby}\disj{}\synt{panda}$ when one of the sentences of the NLI problem asserts the existence of a baby panda: \emph{Two baby pandas are playing} (\sickid{5435}).%
\footnote{The sentences or problems drawn from the \sickd{} dataset \cite{marelli-etal-2014-sick} are supplemented with the problem IDs.
}
Both filters concerning semantic consistency with KB or sentences are used to weed out pseudo-knowledge.

Another adopted criterion for the best T-sets is an \keycon{impact on accuracy} on the training data, which is calculated as a difference between the number of solved and unsolved problems in the training data when a T-set is adopted.
T-sets with no positive impact on accuracy will be ignored as it doesn't contribute to the performance.
This criterion can be broadly related to \eqref{A1}, where a T-set is supposed to conform to the gold labels. 
For example, if a T-set $\{ \synt{boy}\subseq{}\synt{young}, \synt{hedgehog}\subseq{}\synt{small}\}$ helps to solve two new problems but unsolves at least two previously solved ones, then it won't be learned.
For the same NLI problem, a T-set with the highest impact will be preferred over others.   
The motivation behind this criterion is to favor the relations that boost accuracy on the training data.

Despite introduced filters and comparison orders for T-sets, some problems can still have more than one best T-sets.
In such cases, we take into account the \keycon{number of terms} in T-sets (by counting occurrences of atomic terms).
At the end we opt for the T-set with the smallest number of terms.
This decision somewhat complies with \eqref{A2}. 

\section{Experiments}
\label{sect:exp}

We design experiments to evaluate learning as abduction for NLI.
First, we implement the abductive learning for LangPro \cite{abzianidze-2017-langpro}.
The implementation takes advantage of Prolog's virtue of satisfying goals in the forward and backward fashion and uses it to apply closure rules in backwards during the search for B-sets.
This leaves the inventory of tableau rules intact.%
\footnote{The code is available at \url{https://github.com/kovvalsky/LangPro}}
The workflow of the abductive learning is depicted in \autoref{fig:learning_pipe}.

Searching for the best T-set is an NP-hard problem.%
\footnote{The NP-complete set cover problem can be reduced to it.}
To make the implementation efficient \eqref{q1}, we significantly reduce a space of T-sets by considering only those T-sets that coincide with B-sets.
In other words, we require existence of the shared B-sets across all open branches.
The example in \autoref{proof:abd_tree} doesn't have such T-set as different B-sets close the open branches.
If a tableau has a single open branch, its B-sets are automatically T-sets. 

To test the learning algorithm, we use the \sickd{} dataset \cite{marelli-etal-2014-sick} for three reasons.
First, compositional lexical knowledge involved in the dataset is suitable for the abductive learning.
Second, it is large enough (up to 10K problems) to support learning from a training part and evaluation on an unseen part.
Third, \sickd{} has been used for evaluating logic-based NLI systems, including LangPro, and this allows comparison to the existing results.
Our data partition follows the SemEval-14 task-1 \cite{marelli-etal-2014-semeval}:
\sickd{}-train\&trial (4500 + 500 problems) is a training data and \sickd{}-test (4927 problems) a hidden test data.
The error analysis is conducted only on the training data.
To choose optimal learning parameters (see \autoref{subsect:best}) and measure impact of the filters and knowledge resources, we run the learning algorithm with stratified three-fold cross-validation (CV) on the training data.
The stratified version is due to a skewed distribution of gold labels in \sickd{}.
We opt for three-fold CV as it better reflects 1:1 ratio between \sickd{}-train\&trial and -test sizes.

Logic-based NLI systems using logical forms from syntactic trees often employ output of several parsers to increase the quality of logical forms \cite{abzianidze-2015-tableau,beltagy-etal-2016-representing,martinez-gomez-etal-2017-demand,yanaka-etal-2018-acquisition}.
In our CV experiments, we use the re-banked C\&C CCG parser \cite{clark-curran-2007-wide,honnibal-etal-2010-rebanking}.
The ensemble of parsers, used for evaluation on \sickd{}-test, additionally includes EasyCCG \cite{lewis-steedman-2014-ccg} and \mbox{DepCCG} \cite{yoshikawa-etal-2017-ccg} with standard models.

\section{Results}
\label{sect:res}

Initially we run LangPro with WordNet \cite{Miller:1995} relations and all filters enabled (see \autoref{subsect:best}). 
The results in \autoref{tab:3fcv} lead to several findings.
The abductive learning does help to improve accuracy.
LangPro without abduction (with 800 rule applications) gets 81.7\% accuracy on \sickd-train\&test while with abduction it obtains 82.9\% on average over unseen parts of CV on \sickd-train\&test.  
Differences between average accuracies on training and test parts show that overfitting during training is moderate.    
When the prover is limited to 50 rule app., accuracy drops only to 82.64\%.
However, the entire CV takes almost 10 times less CPU time for 50 rule app. compared to 800.
These results answer \eqref{q1} by rendering the abductive learning as a computationally feasible learning method.

We conduct ablation experiments to verify the contributions of the filters.
\autoref{tab:3fcv} shows that filters concerning semantic consistency and syntactic comparability together have little impact on accuracy (.32\%), but they contribute to efficiency by halving CPU time.
This means that other filters like impact on accuracy and term size greatly contribute to preventing pseudo-knowledge.
When relations in T-sets are restricted to atomic terms (i.e., terms of length 1), accuracy is almost unchanged (-0.16\%) while efficiency slightly increases.
The increase in efficiency is clear as fewer T-sets will be considered during learning.
Little change in accuracy means that mostly relations over atomic terms generalize well over the unseen part.

We also tested whether the learned relations can compensate for the WordNet relations \eqref{q3}.
The results show that WordNet relations still contribute to high accuracy as their exclusion drops accuracy by 2.44\%.
\citet{abzianidze-2015-tableau} uses hand-crafted KB of $\sim$30 lexical relations collected from \sickd{}-train\&trial.
When this KB is add, only six more problems on average (0.36\%) is classified correctly.   

To answer \eqref{q4}, we compare LangPro with and without abduction on unseen \sickd{}-test.
The results in \autoref{tab:evaluation} show that the abductive learning consistently increases accuracy regardless of the used CCG parser.
This also means that the abductive learning generalizes across CCG parsers.
When predictions of three LangPro versions with C\&C, EasyCCG and DepCCG parsers are aggregated, accuracy gain from abduction still remains (+1.36\%, additional 67 problems solved). 
It is worth noting that while abduction increases the overall accuracy, almost perfect precision (\textgreater97.6\%) of LangPro decreases only to 94.3\%.
We argue that this is an important virtue of the abductive learning from a logic perspective since logic-based NLI systems are expected to have highly reliable proofs.

\begin{table}[t]
\hspace*{-1mm}
\scalebox{.85}{
\begin{tabular}{@{} l @{} c @{~~} c @{} r @{}}
\toprule
\multirow{2}{*}{\tabul{LangPro + Abductive learning:
                       \\{\small All filters + WordNet}}}
    & \multirow{2}{*}{\tabuc{Train\\av.\,acc\%}}
    & \multirow{2}{*}{\tabuc{Test\\av.\,acc\%}}
    & \multirow{2}{*}{\tabuc{CPU\\time}}
\\
 & &
\\\cmidrule{2-4} 
\kern4mm {max 800 rule applications} & 89.02 & 82.90 & 2041
 \\
\kern4mm {max 50 rule applications} & 88.57 & 82.64 & 220
\\\midrule
{Ablation with max 50 rule app.} & $\Delta$ & $\Delta$ & $\Delta$~~ 
\\\arrayrulecolor{gray}
\midrule
\arrayrulecolor{black}
\kern4mm$-$ CT, CwS, CwKB & +1.39 & -0.32 & +115\% 
\\
\kern4mm$-$ terms of length 2 \& 3 & -2.68 & -0.16 & -17\% 
\\\arrayrulecolor{gray!50}
\midrule
\arrayrulecolor{black}
\kern4mm$-$ WN rels: ant,hyp,der,sim & -0.98 & -2.44 & -11\% 
\\
\kern4mm$+$ Hand-crafted \WritingHand{} KB & +0.04 & +0.36 & +1\% 
\\\bottomrule
\end{tabular}
}
\vspace{-1mm}
\caption{Results of the CV on \sickd{}-train\&trial (majority baseline = 56.4\%).
Ablation experiments disable filters for comparable terms (CT) and consistencies with sentences (CwS) and KB (CwKB).  
CPU time (for 2.5 GHz) is measured in minutes for the entire CV. 
}
\label{tab:3fcv}
\end{table}

\section{Error Analysis}
\label{sect:analysis}

In total 312 relations were learned with abduction from \sickd{}-train\&trial based on a single parser. 
\autoref{tab:learned_knowledge} lists some of the learned relations.
We classify relations into four groups based on whether they are mostly \emph{correct}, \emph{wrong}, \emph{reversed} version of a correct relation, and highly \emph{context-dependent}.

Despite having a sequence of filters, substantial pseudo-knowledge (29\%) still leaked during the learning.
One of the main reasons for this is a strong learning bias towards minimal explanations which often leads to ignoring context.
This way, from \sickid{9624} \emph{\ldots is looking toward the stars\ldots} entailing \emph{\ldots is looking toward the sky\ldots}, $\synt{star}\subseq{}\synt{sky}$ relation was wrongly learned.
Additionally, learning $\synt{in the dark}\subseq{}\synt{at night}$ is preferred to currently learned $\synt{in}\subseq{}\synt{at}$.
This is a common drawback of pure logic-based approaches which is induced by a general principle, called a \emph{rule of replacement}, which licenses replacement of equivalent terms.

A number of incorrect learned relations are conditioned by noisy gold labels of \sickd{} \cite{kalouli-etal-2017-correcting}.
$\synt{dog}\subseq{}\synt{bull dog}$ was learned due to \sickid{2608}, \emph{A monkey is brushing the dog} contradicting \emph{The monkey is not brushing a bull dog}; $\synt{person}\subseq{}\synt{man}$ is due to \sickid{4680}, \emph{Someone is drilling a hole in a strip of wood with a power drill} entailing \emph{A man is drilling a hole in a piece of wood}.

\begin{table}[t]
\hspace*{-1mm}\scalebox{.9}{
\begin{tabular}{@{}l @{\kern-8mm} r@{}}\toprule[1.5pt]
\kern10mm Correct (26.6\%) & Wrong (28.8\%) \kern10mm \\
\midrule
$\synt{add}$ \cntxt{X to Y}$\disj{}\synt{remove}$ \cntxt{X from Y} 
    & \cntxt{the} $\synt{blonde}$ \cntxt{girl}$\disj{}$\cntxt{a} $\synt{little}$ \cntxt{girl}
\\
\cntxt{a X} $\synt{lie down}\disj{}$\cntxt{a X} $\synt{run around}$ 
    & $\synt{aim}$ \cntxt{a gun} $\subseq{}\,\synt{draw}$ \cntxt{a gun}
\\
\cntxt{perform} $\synt{acrobatics}\subseq{}$\cntxt{perform a} $\synt{trick}$
    & $\synt{ride}$ \cntxt{in X}$\disj{}\synt{get}$ \cntxt{out of X}
\\\midrule[1.5pt]
\kern10mm Reversed (28.9\%) & Contextual (15.7\%) \kern10mm
\\\midrule
$\synt{have}$ \cntxt{lunch} $\subseq{}\synt{eat}$ 
    & \cntxt{hang on a} $\synt{cord}\subseq{}$\cntxt{hang on a} $\synt{rope}$
\\
$\synt{look}$ \cntxt{at X} $\subseq{}\synt{stare}$ \cntxt{at X} 
    & \cntxt{man in a} $\synt{cap}\subseq{}$\cntxt{man in a} $\synt{hat}$
\\
\cntxt{prepare some} $\synt{food}\subseq{}$\cntxt{prepare a} $\synt{meal}$ 
    & $\synt{in}$ \cntxt{the dark} $\subseq{}\,\synt{at}$ \cntxt{night}
\\\midrule
\end{tabular}}
\vspace{-2mm}
\caption{Examples of learned lexical relations.
The terms of the relations are in boldface while padded gray contexts come from the source \sickd{} problems.
The relations are manually assessed by the author outside context of \sickd{} problems.
}
\label{tab:learned_knowledge}
\end{table}

\section{Related Work \& Comparison}
\label{sect:rel}

The closest work to ours, to the best of our knowledge, represents \citet{yanaka-etal-2018-acquisition}.
They use abduction for a logic-based NLI system to automatically acquire phrase correspondences from labeled NLI problems.
Their method, called P2P, converts logical formulas into graphs and carries out subgraph matching with variable unification.
Our work differs from theirs in four aspects: 
(i) employed formal logics and proof procedures essentially differ from each other.%
\footnote{\citet{yanaka-etal-2018-acquisition} employs higher-order logic most fragment of which is first-order while most of the logical forms used by Natural Tableau are higher-order.
Their system is based on natural deduction while ours on semantic tableau. 
}
(ii) Converting formulas into graphs and matching subgraphs is external to their theorem proving while in our approach the abductive learning is the theorem proving run backwards.
(iii) P2P abstracts from term comparability constraint and learns relations across word classes.
It also reduces lexical relations to smaller axioms.%
\footnote{For example, P2P captures \emph{cut} entails \emph{chop down} by learning $\forall x (\text{cut}(x)\!\to\!\text{chop}(x))$ and $\forall x (\text{cut}(x)\!\to\!\text{down}(x))$.}
In total P2P extracts 9445 axioms from \sickd{}-train\&trial compared to 312 relations by our abductive learning. 
(iv) P2P scarifies a substantial amount of precision (12.9\%) to gain 1.2\% of accuracy while our abductive learning achieves more gain in accuracy with much less drop in precision.

\bgroup
\begin{table}[t]
\hspace*{-2mm}
\scalebox{.8}{
\begin{tabular}{@{}l @{\kern-1mm}c @{\kern1mm} c @{~} c @{\kern0mm} c @{~~} c @{~~~} c @{~~~} c @{}}
\toprule
~~ System & \kern-4mmParsers & \kern-1mmLearn & \kern-3mmMLc & KB\&Res. & P\% & R\% & A\%
\\\midrule
\ding{229}LangPro & C & $-$ & $-$ & WN & 97.8 & 58.0 & 81.3 
\\
\ding{229}LangPro & C & Abd & $-$ & WN & 94.9 & 63.4 & 82.7 
\\
\ding{229}LangPro & E & $-$ & $-$ & WN & 97.7 & 57.7 & 81.1 
\\
\ding{229}LangPro & E & Abd & $-$ & WN & 94.9 & 63.0 & 82.5 
\\
\ding{229}LangPro & D & $-$ & $-$ & WN & 97.8 & 59.2 & 81.8 
\\
\ding{229}LangPro & D & Abd & $-$ & WN & 94.8 & 64.3 & 83.0 
\\
\ding{229}LangPro & CDE & $-$ & $-$ & WN & 97.6 & 62.2 & 83.0 
\\
\ding{229}\textbf{LangPro} & CDE & Abd & $-$ & WN & 94.3 & \textbf{67.9} & \textbf{84.4} 
\\
LangPro \citeyear{abzianidze-2015-tableau}& \phantom{D}CE & $-$ & $-$ & \kern-3mmWN,\WritingHand{}KB & \textbf{98.0} & 58.1 & 81.4
\\\arrayrulecolor{gray!50}
\midrule[2pt]
\arrayrulecolor{black}
MG et al. \citeyear{martinez-gomez-etal-2017-demand} 
    & \phantom{D}CE & ~W2W & $-$ & WN,VO & \textbf{97.1} & 63.6 & 83.1 
\\
\citeauthor{yanaka-etal-2018-acquisition} & CDE 
    & W2W,P2P\kern-4mm & $-$ & WN,VO & 84.2 & \textbf{77.3} & \textbf{84.3}
\\\arrayrulecolor{gray!50}
\midrule[1pt]
\arrayrulecolor{black}
\ding{72}\citeauthor{bjerva-etal-2014-meaning} & C & $-$ & SVM & WN,PP & 93.6 & 60.6 & \textbf{81.6}
\\
\citeauthor{pavlick-etal-2015-adding} & C & $-$ & $-$ 
    & \multicolumn{3}{l}{WN,PP$^{+}$} & 78.4
\\\arrayrulecolor{gray!50}
\midrule[1pt]
\arrayrulecolor{black}
\ding{72}\citeauthor{beltagy-etal-2014-utexas} & C & $-$ & SVM & WN & 97.9 & 38.7 & 73.2
\\
\citeauthor{beltagy-erk-2015-proper} & C & $-$ & SVM & WN & & & 76.5
\\
\citeauthor{beltagy-etal-2016-representing} & \kern-3mmCE & \kern-3mmRob.Res.\kern-1mm & SVM 
    & \multicolumn{3}{l}{WN,PP,Dist,\WritingHand{}Rules} & \textbf{85.1}
\\\arrayrulecolor{gray!50}
\midrule[2pt]
\arrayrulecolor{black}
\citet{hu2020monalog} & CE\kern-2mm & $-$ & $-$ & WN & 83.8 & 70.7 & 77.2
\\
\multicolumn{7}{l}{\kern5mm $+$ BERT \cite{devlin-etal-2019-bertfrom}} & \textbf{85.4}
\\\arrayrulecolor{gray!50}
\midrule[2pt]
\arrayrulecolor{black}
\multicolumn{7}{l}{\kern-2mm\ding{72}\citeauthor{lai-hockenmaier-2014-illinois} (winner of the SemEval task)} & \textbf{84.6}
\\
\multicolumn{7}{l}{\kern-2mm\citeauthor{yin-schutze-2017-task} DL with GRU \& Attentive Pooling} & \textbf{87.1}
\\\bottomrule
\end{tabular}
}
\vspace{-1mm}
\caption{Comparison of LangPro$_{800}$+Abduction and other logic-based systems on \sickd{}-test.
Some results are not directly comparable as the systems use different KB, resources, CCG parsers, or even employ a machine learning classifier (MLc).
Systems are grouped based on their characteristic approaches to NLI.
The last two systems are not based on logic.
A list of abbreviations: current work (\ding{229}), SemEval-14 task-1 participants (\ding{72}), C\&C parser (C), EasyCCG (E), DepCCG (D), PPDB (PP,  \citealp{ganitkevitch2013ppdb}),
and VerbOcean (VO, \citealp{chklovski-pantel-2004-verbocean}).  
}
\label{tab:evaluation}
\end{table}
\egroup

There have been several logic-based NLI systems evaluated on \sickd{}.
We believe \autoref{tab:evaluation} lists all (but not only) of those systems along with their scores.
The research line by \citet{mineshima-etal-2015-higher,martinez-gomez-etal-2017-demand,yanaka-etal-2018-acquisition} was already described while comparing their approach to ours.
The work by \citet{bjerva-etal-2014-meaning,pavlick-etal-2015-adding} employ Boxer \cite{Bos2008STEP2} to obtain first-order logic formulas from sentences and use an SVM classifier on top of Nutcracker \cite{BosMarkert2005EMNLP}, which reasons using off-the-shelf theorem prover and model builder.
\citet{beltagy-etal-2014-utexas,beltagy-etal-2016-representing} also uses Boxer to get first-order logic formulas but employs probabilistic logic inference in Markov Logic Networks.
To hit the high score on \sickd{}, they combine multiple components including distributional semantics, a set of hand-crafted rules, resolution-based on-fly generation of inference rules, and an SVM classifier as the final predictor.
\citet{hu-etal-2019-natural} use a lightweight system, called MonaLog, based on monotonicity reasoning.
It is further combined with BERT \cite{devlin-etal-2019-bertfrom} to reclassify problems that were predicted by MonaLog as neutral.

Abductive reasoning was already employed by \citet{AAAI05RainaNgManning} at the first Recognizing Textual Entailment challenge \cite{Dagan:2005}.
They used resolution method and a learned cost model to select the cheapest set of assumptions supporting the entailment.
\citet{Hobbs93interpretationas} uses weighted abduction to model text interpretation as the minimal explanation of why text would be true.
The title of the current paper is inspired by this work.

\section{Conclusion}
\label{sect:con}

\autoref{tab:evaluation} shows that the abductive learning component is crucial for logic-based reasoning systems to achieve competitive results.
We have implemented and showed that learning as abduction works successfully for tableau theorem prover the theorem prover to learn lexical relations from data.
Our findings answer the predefined research question as follows.
\eqref{q1} Implementing abduction as backwards theorem proving represents a computationally feasible approach for data-driven learning.
This was achieved by reducing the explanation space: considering only those T-sets that are shared by all open branches and applying several filters to them.     
\eqref{q2} Pseudo-knowledge is partially prevented with a sequence of filters and comparison criteria.
Abductive bias towards minimality often leads to relations that require additional context.
Overall, pseudo-knowledge doesn't harm high precision of the theorem proving.  
\eqref{q3} Despite knowledge learned from data, the lexical relations extracted from WordNet are crucial to reach the state-of-the-art results.
\eqref{q4} Abductive learning consistently increases the accuracy score of the prover regardless of using different parsers individually or in ensemble.

For future work it is interesting to explore the ways that consider larger explanation space and are not strictly preferring short phrases to longer ones.
The latter will allow relations with more context. 


\section*{Acknowledgments}
I would like to thank three anonymous reviewers for their valuable comments and the CIT of the University of Groningen for providing access to the Peregrine HPC cluster.
This work was supported by the NWO-VICI grant (288-89-003) while I was at the University of Groningen and by the European Research Council (ERC) under the European Unions Horizon 2020 research and innovation programme (grant agreement No.\,742204) since I joined Utrecht University.

\bibliography{acl2020}

\begin{thebibliography}{41}
\expandafter\ifx\csname natexlab\endcsname\relax\def\natexlab#1{#1}\fi

\bibitem[{Abzianidze(2015)}]{abzianidze-2015-tableau}
Lasha Abzianidze. 2015.
\newblock \href {https://doi.org/10.18653/v1/D15-1296} {A tableau prover for
  natural logic and language}.
\newblock In \emph{Proceedings of the 2015 Conference on Empirical Methods in
  Natural Language Processing}, pages 2492--2502, Lisbon, Portugal. Association
  for Computational Linguistics.

\bibitem[{Abzianidze(2017{\natexlab{a}})}]{abzianidze-2017-langpro}
Lasha Abzianidze. 2017{\natexlab{a}}.
\newblock \href {https://doi.org/10.18653/v1/D17-2020} {{L}ang{P}ro: Natural
  language theorem prover}.
\newblock In \emph{Proceedings of the 2017 Conference on Empirical Methods in
  Natural Language Processing: System Demonstrations}, pages 115--120,
  Copenhagen, Denmark. Association for Computational Linguistics.

\bibitem[{Abzianidze(2017{\natexlab{b}})}]{abzianidzethesis}
Lasha Abzianidze. 2017{\natexlab{b}}.
\newblock \emph{A natural proof system for natural language}.
\newblock Ph.D. thesis, Tilburg University.

\bibitem[{Beltagy et~al.(2016)Beltagy, Roller, Cheng, Erk, and
  Mooney}]{beltagy-etal-2016-representing}
I.~Beltagy, Stephen Roller, Pengxiang Cheng, Katrin Erk, and Raymond~J. Mooney.
  2016.
\newblock \href {https://doi.org/10.1162/COLI_a_00266} {Representing meaning
  with a combination of logical and distributional models}.
\newblock \emph{Computational Linguistics}, 42(4):763--808.

\bibitem[{Beltagy and Erk(2015)}]{beltagy-erk-2015-proper}
Islam Beltagy and Katrin Erk. 2015.
\newblock \href {https://www.aclweb.org/anthology/W15-0119} {On the proper
  treatment of quantifiers in probabilistic logic semantics}.
\newblock In \emph{Proceedings of the 11th International Conference on
  Computational Semantics}, pages 140--150, London, UK. Association for
  Computational Linguistics.

\bibitem[{Beltagy et~al.(2014)Beltagy, Roller, Boleda, Erk, and
  Mooney}]{beltagy-etal-2014-utexas}
Islam Beltagy, Stephen Roller, Gemma Boleda, Katrin Erk, and Raymond Mooney.
  2014.
\newblock \href {https://doi.org/10.3115/v1/S14-2141} {{UT}exas: Natural
  language semantics using distributional semantics and probabilistic logic}.
\newblock In \emph{Proceedings of the 8th International Workshop on Semantic
  Evaluation ({S}em{E}val 2014)}, pages 796--801, Dublin, Ireland. Association
  for Computational Linguistics.

\bibitem[{van Benthem(2008)}]{Benthem:NatLog:2008}
Johan van Benthem. 2008.
\newblock A brief history of natural logic.
\newblock In \emph{Technical Report PP-2008-05}. {Institute for Logic, Language
  \& Computation}.

\bibitem[{Bjerva et~al.(2014)Bjerva, Bos, van~der Goot, and
  Nissim}]{bjerva-etal-2014-meaning}
Johannes Bjerva, Johan Bos, Rob van~der Goot, and Malvina Nissim. 2014.
\newblock \href {https://doi.org/10.3115/v1/S14-2114} {The meaning factory:
  Formal semantics for recognizing textual entailment and determining semantic
  similarity}.
\newblock In \emph{Proceedings of the 8th International Workshop on Semantic
  Evaluation ({S}em{E}val 2014)}, pages 642--646, Dublin, Ireland. Association
  for Computational Linguistics.

\bibitem[{Bos(2008)}]{Bos2008STEP2}
Johan Bos. 2008.
\newblock Wide-coverage semantic analysis with boxer.
\newblock In \emph{Semantics in Text Processing. STEP 2008 Conference
  Proceedings}, Research in Computational Semantics, pages 277--286. College
  Publications.

\bibitem[{Bos and Markert(2005)}]{BosMarkert2005EMNLP}
Johan Bos and Katja Markert. 2005.
\newblock Recognising textual entailment with logical inference.
\newblock In \emph{Proceedings of the 2005 Conference on Empirical Methods in
  Natural Language Processing (EMNLP 2005)}, pages 628--635.

\bibitem[{Chklovski and Pantel(2004)}]{chklovski-pantel-2004-verbocean}
Timothy Chklovski and Patrick Pantel. 2004.
\newblock \href {https://www.aclweb.org/anthology/W04-3205} {{V}erb{O}cean:
  Mining the web for fine-grained semantic verb relations}.
\newblock In \emph{Proceedings of the 2004 Conference on Empirical Methods in
  Natural Language Processing}, pages 33--40, Barcelona, Spain. Association for
  Computational Linguistics.

\bibitem[{Clark and Curran(2007)}]{clark-curran-2007-wide}
Stephen Clark and James~R. Curran. 2007.
\newblock \href {https://doi.org/10.1162/coli.2007.33.4.493} {Wide-coverage
  efficient statistical parsing with {CCG} and log-linear models}.
\newblock \emph{Computational Linguistics}, 33(4):493--552.

\bibitem[{Cooper et~al.(1996)Cooper, Crouch, Eijck, Fox, Genabith, Jaspars,
  Kamp, Milward, Pinkal, Poesio, Pulman, Briscoe, Maier, and Konrad}]{fracas}
Robin Cooper, Dick Crouch, Jan~Van Eijck, Chris Fox, Josef~Van Genabith, Jan
  Jaspars, Hans Kamp, David Milward, Manfred Pinkal, Massimo Poesio, Steve
  Pulman, Ted Briscoe, Holger Maier, and Karsten Konrad. 1996.
\newblock \emph{FraCaS: A Framework for Computational Semantics}.
\newblock Deliverable D16.

\bibitem[{Dagan et~al.(2005)Dagan, Glickman, and Magnini}]{Dagan:2005}
Ido Dagan, Oren Glickman, and Bernardo Magnini. 2005.
\newblock The pascal recognising textual entailment challenge.
\newblock In \emph{Proceedings of the PASCAL Challenges Workshop on Recognising
  Textual Entailment}.

\bibitem[{Dagan et~al.(2013)Dagan, Roth, Sammons, and Zanzotto}]{rteBook:2013}
Ido Dagan, Dan Roth, Mark Sammons, and Fabio~Massimo Zanzotto. 2013.
\newblock \emph{Recognizing Textual Entailment: Models and Applications}.
\newblock Synthesis Lectures on Human Language Technologies. Morgan {\&}
  Claypool Publishers.

\bibitem[{Devlin et~al.(2019)Devlin, Chang, Lee, and
  Toutanova}]{devlin-etal-2019-bertfrom}
Jacob Devlin, Ming-Wei Chang, Kenton Lee, and Kristina Toutanova. 2019.
\newblock \href {https://doi.org/10.18653/v1/N19-1423} {{BERT}: Pre-training of
  deep bidirectional transformers for language understanding}.
\newblock In \emph{Proceedings of the 2019 Conference of the North {A}merican
  Chapter of the Association for Computational Linguistics: Human Language
  Technologies, Volume 1 (Long and Short Papers)}, pages 4171--4186,
  Minneapolis, Minnesota. Association for Computational Linguistics.

\bibitem[{Ganitkevitch et~al.(2013)Ganitkevitch, {Van Durme}, and
  Callison-Burch}]{ganitkevitch2013ppdb}
Juri Ganitkevitch, Benjamin {Van Durme}, and Chris Callison-Burch. 2013.
\newblock \href {http://cs.jhu.edu/~ccb/publications/ppdb.pdf} {{PPDB}: The
  paraphrase database}.
\newblock In \emph{Proceedings of NAACL-HLT}, pages 758--764, Atlanta, Georgia.
  Association for Computational Linguistics.

\bibitem[{Glockner et~al.(2018)Glockner, Shwartz, and
  Goldberg}]{glockner-etal-2018-breaking}
Max Glockner, Vered Shwartz, and Yoav Goldberg. 2018.
\newblock \href {https://doi.org/10.18653/v1/P18-2103} {Breaking {NLI} systems
  with sentences that require simple lexical inferences}.
\newblock In \emph{Proceedings of the 56th Annual Meeting of the Association
  for Computational Linguistics (Volume 2: Short Papers)}, pages 650--655,
  Melbourne, Australia. Association for Computational Linguistics.

\bibitem[{Gururangan et~al.(2018)Gururangan, Swayamdipta, Levy, Schwartz,
  Bowman, and Smith}]{gururangan-etal-2018-annotation}
Suchin Gururangan, Swabha Swayamdipta, Omer Levy, Roy Schwartz, Samuel Bowman,
  and Noah~A. Smith. 2018.
\newblock \href {https://doi.org/10.18653/v1/N18-2017} {Annotation artifacts in
  natural language inference data}.
\newblock In \emph{Proceedings of the 2018 Conference of the North {A}merican
  Chapter of the Association for Computational Linguistics: Human Language
  Technologies, Volume 2 (Short Papers)}, pages 107--112, New Orleans,
  Louisiana. Association for Computational Linguistics.

\bibitem[{Hobbs et~al.(1993)Hobbs, Stickel, and
  Martin}]{Hobbs93interpretationas}
Jerry~R. Hobbs, Mark Stickel, and Paul Martin. 1993.
\newblock Interpretation as abduction.
\newblock \emph{Artificial Intelligence}, 63:69--142.

\bibitem[{Honnibal et~al.(2010)Honnibal, Curran, and
  Bos}]{honnibal-etal-2010-rebanking}
Matthew Honnibal, James~R. Curran, and Johan Bos. 2010.
\newblock \href {https://www.aclweb.org/anthology/P10-1022} {Rebanking
  {CCG}bank for improved {NP} interpretation}.
\newblock In \emph{Proceedings of the 48th Annual Meeting of the Association
  for Computational Linguistics}, pages 207--215, Uppsala, Sweden. Association
  for Computational Linguistics.

\bibitem[{Hu et~al.(2019)Hu, Chen, and Moss}]{hu-etal-2019-natural}
Hai Hu, Qi~Chen, and Larry Moss. 2019.
\newblock \href {https://www.aclweb.org/anthology/W19-0502} {Natural language
  inference with monotonicity}.
\newblock In \emph{Proceedings of the 13th International Conference on
  Computational Semantics - Short Papers}, pages 8--15, Gothenburg, Sweden.
  Association for Computational Linguistics.

\bibitem[{Hu et~al.(2020)Hu, Chen, Richardson, Mukherjee, Moss, and
  K{\"u}bler}]{hu2020monalog}
Hai Hu, Qi~Chen, Kyle Richardson, Atreyee Mukherjee, Lawrence~S Moss, and
  Sandra K{\"u}bler. 2020.
\newblock Monalog: a lightweight system for natural language inference based on
  monotonicity.
\newblock \emph{Proceedings of the Society for Computation in Linguistics},
  3(1):319--329.

\bibitem[{Kalouli et~al.(2017)Kalouli, de~Paiva, and
  Real}]{kalouli-etal-2017-correcting}
Aikaterini-Lida Kalouli, Valeria de~Paiva, and Livy Real. 2017.
\newblock \href {https://www.aclweb.org/anthology/W17-7205} {Correcting
  contradictions}.
\newblock In \emph{Proceedings of the Computing Natural Language Inference
  Workshop}.

\bibitem[{Lai and Hockenmaier(2014)}]{lai-hockenmaier-2014-illinois}
Alice Lai and Julia Hockenmaier. 2014.
\newblock \href {https://doi.org/10.3115/v1/S14-2055} {{I}llinois-{LH}: A
  denotational and distributional approach to semantics}.
\newblock In \emph{Proceedings of the 8th International Workshop on Semantic
  Evaluation ({S}em{E}val 2014)}, pages 329--334, Dublin, Ireland. Association
  for Computational Linguistics.

\bibitem[{Lewis and Steedman(2014)}]{lewis-steedman-2014-ccg}
Mike Lewis and Mark Steedman. 2014.
\newblock \href {https://doi.org/10.3115/v1/D14-1107} {{A}* {CCG} parsing with
  a supertag-factored model}.
\newblock In \emph{Proceedings of the 2014 Conference on Empirical Methods in
  Natural Language Processing ({EMNLP})}, pages 990--1000, Doha, Qatar.
  Association for Computational Linguistics.

\bibitem[{Marelli et~al.(2014{\natexlab{a}})Marelli, Bentivogli, Baroni,
  Bernardi, Menini, and Zamparelli}]{marelli-etal-2014-semeval}
Marco Marelli, Luisa Bentivogli, Marco Baroni, Raffaella Bernardi, Stefano
  Menini, and Roberto Zamparelli. 2014{\natexlab{a}}.
\newblock \href {https://doi.org/10.3115/v1/S14-2001} {{S}em{E}val-2014 task 1:
  Evaluation of compositional distributional semantic models on full sentences
  through semantic relatedness and textual entailment}.
\newblock In \emph{Proceedings of the 8th International Workshop on Semantic
  Evaluation ({S}em{E}val 2014)}, pages 1--8, Dublin, Ireland. Association for
  Computational Linguistics.

\bibitem[{Marelli et~al.(2014{\natexlab{b}})Marelli, Menini, Baroni,
  Bentivogli, Bernardi, and Zamparelli}]{marelli-etal-2014-sick}
Marco Marelli, Stefano Menini, Marco Baroni, Luisa Bentivogli, Raffaella
  Bernardi, and Roberto Zamparelli. 2014{\natexlab{b}}.
\newblock \href
  {http://www.lrec-conf.org/proceedings/lrec2014/pdf/363_Paper.pdf} {A {SICK}
  cure for the evaluation of compositional distributional semantic models}.
\newblock In \emph{Proceedings of the Ninth International Conference on
  Language Resources and Evaluation ({LREC}-2014)}, pages 216--223, Reykjavik,
  Iceland. European Languages Resources Association (ELRA).

\bibitem[{Mart{\'\i}nez-G{\'o}mez et~al.(2017)Mart{\'\i}nez-G{\'o}mez,
  Mineshima, Miyao, and Bekki}]{martinez-gomez-etal-2017-demand}
Pascual Mart{\'\i}nez-G{\'o}mez, Koji Mineshima, Yusuke Miyao, and Daisuke
  Bekki. 2017.
\newblock \href {https://www.aclweb.org/anthology/E17-1067} {On-demand
  injection of lexical knowledge for recognising textual entailment}.
\newblock In \emph{Proceedings of the 15th Conference of the {E}uropean Chapter
  of the Association for Computational Linguistics: Volume 1, Long Papers},
  pages 710--720, Valencia, Spain. Association for Computational Linguistics.

\bibitem[{Mayer and Pirri(1993)}]{mayer_pirri:93}
Marta~Cialdea Mayer and Fiora Pirri. 1993.
\newblock \href {https://doi.org/10.1093/jigpal/1.1.99} {{First order abduction
  via tableau and sequent calculi}}.
\newblock \emph{Logic Journal of the IGPL}, 1(1):99--117.

\bibitem[{Miller(1995)}]{Miller:1995}
George~A. Miller. 1995.
\newblock Wordnet: A lexical database for english.
\newblock \emph{Communications of the ACM}, 38(11):39--41.

\bibitem[{Mineshima et~al.(2015)Mineshima, Mart{\'\i}nez-G{\'o}mez, Miyao, and
  Bekki}]{mineshima-etal-2015-higher}
Koji Mineshima, Pascual Mart{\'\i}nez-G{\'o}mez, Yusuke Miyao, and Daisuke
  Bekki. 2015.
\newblock \href {https://doi.org/10.18653/v1/D15-1244} {Higher-order logical
  inference with compositional semantics}.
\newblock In \emph{Proceedings of the 2015 Conference on Empirical Methods in
  Natural Language Processing}, pages 2055--2061, Lisbon, Portugal. Association
  for Computational Linguistics.

\bibitem[{Moss(2010)}]{Moss2010}
Lawrence~S. Moss. 2010.
\newblock Natural logic and semantics.
\newblock In Maria Aloni, Harald Bastiaanse, Tikitu de~Jager, and Katrin
  Schulz, editors, \emph{Logic, Language and Meaning: 17th Amsterdam
  Colloquium, Amsterdam, The Netherlands, December 16-18, 2009, Revised
  Selected Papers}, pages 84--93. Springer Berlin Heidelberg, Berlin,
  Heidelberg.

\bibitem[{Muskens(2010)}]{muskens:10}
Reinhard Muskens. 2010.
\newblock An analytic tableau system for natural logic.
\newblock In Maria Aloni, Harald Bastiaanse, Tikitu de~Jager, and Katrin
  Schulz, editors, \emph{Logic, Language and Meaning}, volume 6042 of
  \emph{Lecture Notes in Computer Science}, pages 104--113. Springer Berlin
  Heidelberg.

\bibitem[{Pavlick et~al.(2015)Pavlick, Bos, Nissim, Beller, Van~Durme, and
  Callison-Burch}]{pavlick-etal-2015-adding}
Ellie Pavlick, Johan Bos, Malvina Nissim, Charley Beller, Benjamin Van~Durme,
  and Chris Callison-Burch. 2015.
\newblock \href {https://doi.org/10.3115/v1/P15-1146} {Adding semantics to
  data-driven paraphrasing}.
\newblock In \emph{Proceedings of the 53rd Annual Meeting of the Association
  for Computational Linguistics and the 7th International Joint Conference on
  Natural Language Processing (Volume 1: Long Papers)}, pages 1512--1522,
  Beijing, China. Association for Computational Linguistics.

\bibitem[{Poliak et~al.(2018)Poliak, Naradowsky, Haldar, Rudinger, and
  Van~Durme}]{poliak-etal-2018-hypothesis}
Adam Poliak, Jason Naradowsky, Aparajita Haldar, Rachel Rudinger, and Benjamin
  Van~Durme. 2018.
\newblock \href {https://doi.org/10.18653/v1/S18-2023} {Hypothesis only
  baselines in natural language inference}.
\newblock In \emph{Proceedings of the Seventh Joint Conference on Lexical and
  Computational Semantics}, pages 180--191, New Orleans, Louisiana. Association
  for Computational Linguistics.

\bibitem[{Raina et~al.(2005)Raina, Ng, and Manning}]{AAAI05RainaNgManning}
Rajat Raina, Andrew~Y. Ng, and Christopher~D. Manning. 2005.
\newblock Robust textual inference via learning and abductive reasoning.
\newblock In \emph{Proceedings of AAAI 2005}. AAAI Press.

\bibitem[{Steedman(2000)}]{Steedman:2000}
Mark Steedman. 2000.
\newblock \emph{The Syntactic Process}.
\newblock MIT Press, Cambridge, MA, USA.

\bibitem[{Yanaka et~al.(2018)Yanaka, Mineshima, Mart{\'\i}nez-G{\'o}mez, and
  Bekki}]{yanaka-etal-2018-acquisition}
Hitomi Yanaka, Koji Mineshima, Pascual Mart{\'\i}nez-G{\'o}mez, and Daisuke
  Bekki. 2018.
\newblock \href {https://doi.org/10.18653/v1/N18-1069} {Acquisition of phrase
  correspondences using natural deduction proofs}.
\newblock In \emph{Proceedings of the 2018 Conference of the North {A}merican
  Chapter of the Association for Computational Linguistics: Human Language
  Technologies, Volume 1 (Long Papers)}, pages 756--766, New Orleans,
  Louisiana. Association for Computational Linguistics.

\bibitem[{Yin and Sch{\"u}tze(2017)}]{yin-schutze-2017-task}
Wenpeng Yin and Hinrich Sch{\"u}tze. 2017.
\newblock \href {https://www.aclweb.org/anthology/E17-1066} {Task-specific
  attentive pooling of phrase alignments contributes to sentence matching}.
\newblock In \emph{Proceedings of the 15th Conference of the {E}uropean Chapter
  of the Association for Computational Linguistics: Volume 1, Long Papers},
  pages 699--709, Valencia, Spain. Association for Computational Linguistics.

\bibitem[{Yoshikawa et~al.(2017)Yoshikawa, Noji, and
  Matsumoto}]{yoshikawa-etal-2017-ccg}
Masashi Yoshikawa, Hiroshi Noji, and Yuji Matsumoto. 2017.
\newblock \href {https://doi.org/10.18653/v1/P17-1026} {{A}* {CCG} parsing with
  a supertag and dependency factored model}.
\newblock In \emph{Proceedings of the 55th Annual Meeting of the Association
  for Computational Linguistics (Volume 1: Long Papers)}, pages 277--287,
  Vancouver, Canada. Association for Computational Linguistics.

\end{thebibliography}
\bibliographystyle{acl_natbib}

\end{document}